\documentclass[letterpaper, 10 pt, conference]{ieeeconf}  % Comment this line out if you need a4paper

\IEEEoverridecommandlockouts                              % This command is only needed if 
\AtBeginDocument{%
  \providecommand\BibTeX{{%
    \normalfont B\kern-0.5em{\scshape i\kern-0.25em b}\kern-0.8em\TeX}}}

\usepackage{graphics} 
\usepackage{amsmath} % assumes amsmath package installed
\usepackage{amssymb} 
\usepackage{graphicx}
\usepackage{subfiles}
\usepackage{multirow}
\usepackage{makecell}
\usepackage{soul}
\usepackage{color}
\usepackage{wasysym}
\usepackage{colortbl}
\usepackage[final]{changes}

\usepackage{hyperref}
\usepackage{cleveref}
\usepackage[T1]{fontenc} 

\crefformat{section}{\S#2#1#3} % see manual of cleveref, section 8.2.1
\crefformat{subsection}{\S#2#1#3}
\crefformat{subsubsection}{\S#2#1#3}

\newcommand*\circled[1]{\tikz[baseline=(char.base)]{
            \node[shape=circle,draw,inner sep=1pt] (char) {#1};}}

\title{\LARGE \bf
\textit{Motion Comparator}: Visual Comparison of Robot Motions
}

\author{Yeping Wang$^{1}$, Alexander Peseckis$^{1}$, Zelong Jiang$^{1}$,
and Michael Gleicher$^{1}$% <-this % stops a space
\thanks{$^{1}$ All authors are with the Department of Computer Sciences, University of Wisconsin-Madison, Madison, WI 53706, USA %
{\tt\small [yeping|gleicher]@cs.wisc.edu \newline
 [peseckis|zjiang287]@wisc.edu}}%
\thanks{This work was supported by National Science Foundation award 2007436, Los Alamos National Laboratory and the Department of Energy.}% <-this % stops a space
\vspace{-30mm}
}

\begin{document}

\maketitle
\thispagestyle{empty}
\pagestyle{empty}

%%%%%%%%%%%%%%%%%%%%%%%%%%%%%%%%%%%%%%%%%%%%%%%%%%%%%%%%%%%%%%%%%%%%%%%%%%%%%%%%
\begin{abstract}

Roboticists compare robot motions for tasks such as parameter tuning, troubleshooting, and deciding between possible motions. However, most existing visualization tools are designed for individual motions and lack the features necessary to facilitate robot motion comparison. 
In this paper, we utilize a rigorous design framework to develop \textit{Motion Comparator}, a web-based tool that facilitates the comprehension, comparison, and communication of robot motions. Our design process identified roboticists' needs, articulated design challenges, and provided corresponding strategies. Motion Comparator includes several key features such as multi-view coordination, quaternion visualization, time warping, and comparative designs. To demonstrate the applications of Motion Comparator, we discuss four case studies in which our tool is used for motion selection, troubleshooting, parameter tuning, and motion review.

\end{abstract}

\vspace{-1mm}
\section{Introduction}
Roboticists often employ various quantitative metrics such as duration or average accuracy to evaluate and compare motions. While these metrics provide an overall summary of a motion, they are insufficient for roboticists to \textit{understand} the finer details between them. For example, such metrics do not allow roboticists to explain quantitative findings, identify unusual events, or evaluate motions in a subjective or task-specific way such as assessing their predictability, legibility, or social acceptance. Roboticists need to acquire a comprehensive understanding of each motion before they can decide which motion to deploy. 
Meanwhile, the act of comparing also facilitates the understanding of each individual motion. 
\iffalse
\begin{figure}[t]
  \centering
  \includegraphics[width=3.35in]{figures/teaser.pdf}
  \caption{Screenshot of Motion Comparator. Our tool provides various visualization approaches to support visual comparison of robot motions. We invite readers to try out our web-based system\protect\footnotemark.}
  \label{fig: teaser}
  \vspace{-5mm}
\end{figure}
\fi
% \footnotetext{Link to our tool: \url{https://uwgraphics.github.io/MotionComparator}}

In this paper, we introduce \textit{Motion Comparator}, a web-based system that facilitates the comprehension, comparison, and communication of robot motions. It provides users with different visualizations of robot motions, each designed to help with motion understanding and comparison. The system supports flexible layout and view coordination to allow users to create workflows that support their needs.

We design our system by applying a rigorous visual comparison design framework \cite{gleicher2017considerations} which involves 1) identifying user \textit{needs}, 2) analyzing the comparative \textit{challenges} and \textit{strategies}, and 3) considering the visual \textit{designs} for comparison. 
First, we identify seven motion properties that are of interest to roboticists through a systematic examination of the abstracts of papers published at major venues. 
%First, we identify seven motion properties that are of interest to roboticists through a systematic examination of adjectives and adverbs used to describe robot motions in the abstract of papers published at major venues. 
Then we identify three major \textit{challenges} that make robot motions difficult: temporal length of robot motions, spatial complexity of robot motions, and complexity of relationships between motions. To assist the visualization of temporally long motions, Motion Comparator unfolds time into space by showing traces in Cartesian space, quaternion space, and joint space. Furthermore, Motion Comparator allows users to select, visualize, and compare motion segments. In order to address spatial complexity, Motion Comparator provides multiple coordinated views that allow users to simultaneously inspect a motion using various visualization methods. To mitigate the complex relationship between motions, Motion Comparator utilizes time warping to automatically align two motions.
Motion comparitor employs multiple views based on the basic comparative designs (juxtaposition, superposition, and explicit encoding) in order to satisfy various comparison needs.
% Finally, we incorporate Motion Comparator with three basic comparative designs (juxtaposition, superposition, and explicit encoding) to satisfy various comparison needs. Motion Comparator has five views that are integrated with mechanisms such as multi-view coordination.
The key features of our system are summarized in Table$\,\,\,$\ref{tab:exiting_tools}. We have open-sourced Motion Comparator\footnote{Source code:  \url{https://github.com/uwgraphics/MotionComparator}} and invite readers to use our web-based system\protect\footnotemark.

\footnotetext{Link to our tool: \url{https://pages.graphics.cs.wisc.edu/MotionComparator}}

\added{The main contribution of this paper is a visualization tool for roboticists to understand, compare, and share robot motions. 
In doing this, we make specific contributions including: (1) we provide a literature-based requirements analysis to articulate the visualization needs of robot motions. The analysis informs future tool design and reveals the diversity of these needs, suggesting that a single visual design is unlikely to meet all requirements. (2) We demonstrate how various visualization approaches, such as traces in different spaces and quaternion visualization, can be adapted to motion comparison problems, providing coverage for the range of user needs. 
(3) We show how multiple visualizations can be integrated by adapting multi-view coordination for motion comparison by applying time-warp based alignment and view synchronization.}
%our tool shows how multiple visualizations can be integrated \hl{for comparsion} by enhancing multi-view coordination with approaches such as time-warp based alignment and comparative designs.}

% By incorporating a variety of data visualization methods, such as quaternion visualization and dimensionality reduction, Motion Comparator exemplifies the benefits of applying data visualization principles within the field of robotics. In addition, Motion Comparator highlights the significance of comparative thinking in robot visualization. We believe that future robot visualization systems can benefit from our comparative thinking which includes the comparative needs we identified (\cref{sec: needs}), the comparative challenges we articulated (\cref{sec:challenges}), the comparative strategies we presented (\cref{sec:challenges}), and the comparative designs we used (\cref{sec:design}).}

\added{In the remainder of the paper, we describe how we designed Motion Comparator in \cref{sec:process}. In \cref{sec:views}, we describe Motion Comparator’s multi-view interface, where each view is specialized to visualize a specific aspect of robot motions. The implementation details are provided in \cref{sec:implementation}. Being able to directly parse \texttt{rosbag} files, Motion Comparator is a convenient tool for Robot Operating System (ROS) users. In \cref{sec:case_studies}, we provide four case studies in which Motion Comparator is utilized for motion selection, troubleshooting, parameter tuning, and motion review.}

\deleted{In the remainder of the paper, we review the relevant prior research that motivates this work in \cref{sec:related_works}. Following a problem overview in \cref{sec:overview}, we describe how we designed Motion Comparator by identifying user needs (\cref{sec: needs}), analyzing comparative challenges and strategies (\cref{sec:challenges}), and considering various comparative designs (\cref{sec:design}). 
We then describe Motion Comparator's five views in \cref{sec:views}.
In \cref{sec:implementation}, we describe our web-based implementation that enables roboticists to share motion visualizations with others, facilitating communication and collaboration. Afterward, we provide four case studies in which Motion Comparator is utilized for motion selection, troubleshooting, parameter tuning, and motion review (\cref{sec:case_studies}). 
Finally, we conclude this paper with a discussion of the limitations and future work. }

% Almost all task requires robot motions to be accurate, smooth, and collision-free. Some tasks also impose some additional requirements such as energy efficiency, predictability, legibility, human likeness, or social acceptance. 

\section{Related Works} \label{sec:related_works}
\deleted{This work draws inspiration from prior robot visualization tools and literature on motion visualization and motion visual comparison.}

\definecolor{rowgray}{gray}{0.95}

\begin{table*}[tb]
\caption{Key Features of Motion Comparator in Contrast to Existing Tools}
\label{tab:exiting_tools}
\vspace{-3mm}
\begin{center}
\begin{tabular}{l|ccccccc|cc}
\hline
\rule{0pt}{1.05\normalbaselineskip}%
\multirow{2}{*}{Name} & \multicolumn{7}{c|}{Motion Visualization} & \multicolumn{2}{c}{Motion Comparison}\\[0.5mm]
\cline{2-10}
\rule{0pt}{1.5\normalbaselineskip}%
& \makecell{3D \\ Scene} & \makecell{Time Series\\ Plots} & \makecell{Scrubbable\\ Timeline Bar} & Shareable & \makecell{Position \\ Trace} & \makecell{Quaternion \\ Trace} & \makecell{Joint \\ Trace}  & \makecell{Time \\ Warping} & \makecell{Comparative\\ Designs} \\[1mm]
\hline
 \rowcolor{rowgray}
%\rule{0pt}{1.0\normalbaselineskip}%
 Rviz {\footnotesize(MoveIt Visual Tools)} & \checkmark & & & & \checkmark & & & & \\
 rqt\_plot {\footnotesize(rqt\_multiplot)} & & \checkmark & & & & & & & \\
 \rowcolor{rowgray}
 robot-log-visualizer   & \checkmark &  \checkmark &  \checkmark &  & & & & &\\
 PlotJuggler & & \checkmark & \checkmark & \checkmark & & & & &\\
 \rowcolor{rowgray}
 Webviz  {\footnotesize(Foxglove Studio)} & \checkmark & \checkmark & \checkmark & \checkmark & & & & & \\
 Klamp't & \checkmark & \checkmark & & & \checkmark & & & & \\
 \rowcolor{rowgray}
 Robowflex  & \checkmark & & &  & & & & & \\
 Motion Comparator & \checkmark & \checkmark & \checkmark & \checkmark & \checkmark & \checkmark & \checkmark & \checkmark  & \checkmark \\
\hline
% \multicolumn{10}{l}{\rule{0pt}{1\normalbaselineskip}%
% $^{\dag}$ Webviz Can share the data but not the layout/states of the webpage. } \\
\end{tabular}
\end{center}
\vspace{-7mm}
\end{table*}

\subsection{Robot Visualization Tools}
% Robot motion visualization
%  \cite{szafir2021connecting}

\replaced{Most existing}{Several} robot visualization tools \replaced{focus on}{are available, but they are designed for} single motions.
\replaced{For example, }{A typical visualization tool is }Rviz, a built-in tool in Robot Operating System (ROS)\replaced{,}{ that} visualizes robot states in a simulated 3D world. MoveIt Visual Tool builds upon Rviz to visualize motions generated by the MoveIt motion planner. The tool animates how a robot executes a motion and draws a trace to represent the end-effector path. Aside from them, the ROS ecosystem offers 2D graph tools such as rqt\_plot or rqt\_multiplot to visualize scalar values such as joint velocities.
\replaced{Unlike these tools that focus on real-time visualization,}{However, these ROS-based tools focus on real-time visualization, lacking interactive controls to adjust time in data playback. In contrast,} robot-log-visualizer places emphasis on offline data visualization and offers a scrubbable timeline bar that enables users to navigate through a motion.
PlotJuggler is an emerging time-series data plotting tool with a scrubbable timeline bar and some built-in functionalities such as computing derivatives and integrals. In addition, PlotJuggler offers users the functionality to customize, store, and retrieve layouts\deleted{ according to their preferences}, which is crucial for communication and collaboration. To further facilitate collaborative efforts, Webviz and Foxglove Studio have been developed as web-based applications that allow users to visualize data by simply opening a browser window and navigating to a URL. In addition to these specialized visualization tools, several general toolboxes, such as Klamp't and Robowflex \cite{kingston2022robowflex} also incorporate features to visualize robot states in a 3D environment. The key features of all the aforementioned tools are outlined in Table \ref{tab:exiting_tools} in comparison to Motion Comparator. 
These tools are designed for robot motion visualization and do \textit{not} provide features that facilitate visual comparison of robot motions. Meanwhile, Motion Comparator is built from the ground up to compare robot motions.

% We do not count tools that focus on physics simulation, such as Gazebo, WeBot, CoppliaSim.

\subsection{Motion Visualization} \label{sec: motion_visualization}

In addition to the robot visualization tools, Motion Comparator also draws inspiration from literature on general motion visualization. \deleted{The knowledge from data visualization literature can inform effective robot interface design \cite{szafir2021connecting}.}
% General motion visualization
Coffey \textit{et al.} \cite{coffey2012visualizing} provides a taxonomy of motion visualizations that dissects motions by their space and time dimensions. Informed by their work, Motion Comparator offers functionalities that assist users in comparing the space and time dimensions of robot motions. 
%
% which outlines three design choices for both the time and space dimensions of motions. The time dimension of motions can be either interactively controlled by users, animated, or shown statically. Similarly, the camera that views the space dimension can be interactively controlled by users, automatically controlled, or placed in a stationary position. One of our design rationales is dissecting motions by their time and space dimensions. 
%
\replaced{We also adapt the standard visualization approach that uses a trace to depict an object's movements.}{Previous research has demonstrated that visualizing an object's position over time can be done with a trace of the object in 3D Cartesian space.} Position traces have been employed in the visualization of human motions \cite{reveret20203d, kovar2002motion}, animal movements \cite{ware2006visualizing}, and the motions of the tooltip of a surgical robot \cite{schroeder2012exploratory}. 
\replaced{These}{In addition to illustrating temporal changes in position,} traces have been enhanced with annotation to augment their informational content, such as using chevrons to encode directions \cite{schroeder2012exploratory, ware2006visualizing}, triangles to encode velocities \cite{russig2022tube}, and color to encode grip force applied by a robot end-effector \cite{schroeder2012exploratory}.

% Quaternion visualization
While an object's position change can be visualized using a position trace, it is more challenging to visualize its orientation over time. Several works seek to visualize orientations on position traces using color \cite{hanson1994visualizing} or coordinate frames \cite{juttler1994visualization, patel2023analysis}\replaced{, which can be}{. These methods are either} unintuitive or visually cluttered.  An alternative body of literature represents orientations in quaternions and seeks ways to visualize quaternion traces, \textit{e.g.}, by projecting quaternions to a 3D space \cite{Hanson2006}. Motion Comparator offers both position traces and quaternion traces to visualize the position and orientation changes of an object, \textit{e.g.}, a robot manipulator's end-effector.

Meanwhile, a thread of works focuses on illustrating robot motions using \textit{static} 2D images \cite{rakita2016motion, johansen2023illustrating, cutting2002representing}, which can be readily inserted into academic publications to facilitate scientific communication. While Motion Comparator is an \textit{interactive} system that incorporates several 2D \replaced{and}{or} 3D visualization approaches, users can take representative screenshots of Motion Comparator and incorporate them into academic papers. All figures in this paper are assembled from one or more screenshots of Motion Comparator.

% PCA for dimension reduction \cite{kruger2016efficient}

\subsection{Motion Visual Comparison} \label{sec:visual_comparison}

% Comparison is an important part in robotics. Some of Erdem Bıyık's work

Gleicher \cite{gleicher2017considerations} identifies three basic designs for visual comparison of general objects: 
\textit{1) Juxtaposition designs} place objects next to each other and rely on the viewer’s memory to make connections between objects. Consequently, subtle differences may not be apparent in juxtaposition designs.
\textit{2) Superposition designs} overlay objects in the same place and facilitate the detection of subtle differences. Superposition designs use the viewer's vision for comparison and one of their notable issues is the occlusion or clutter caused by overlapped objects.
\textit{3) Explicit encoding designs} compute relationships between objects and provide visual encoding of the relationships. While relieving viewers from the burden of comparison, explicit encodings require the relationships to be known and viewers may experience a loss of contextual information. Explicit encoding is beneficial in measuring or communicating a relationship but may not be as effective in contextualizing or dissecting a relationship.
Kim \textit{et al.} \cite{kim2017comparison} distinguish spatial and temporal juxtaposition, calling the latter category interchangeable designs as such designs alternate between items temporally in the same space. 
In \cref{sec:design}, we will further discuss these three designs with their benefits and drawbacks. 
To satisfy the various needs of robot motion comparison, Motion Comparator is integrated with all juxtaposition, superposition, and explicit encoding designs.   

\replaced{To facilitate comparison, time warping is often used to temporally align multiple motions}{Time warping is a commonly used technique to facilitate the comparison between time-series data} \cite{vaughan2016comparing}. Gong \textit{et al.} \cite{gong2022motion} use time warping to compare the motion between a teleoperated robot and its human operator. The alignment after time warping can be visualized by connecting corresponding time frames \cite{giorgino2009computing}, using warping curves \cite{giorgino2009computing}, or encoding in colors \cite{gogolou2018comparing}. Motion Comparator also offers time warping functionality and provides built-in loss functions to customize the temporal alignment of two robot motions.

% This paper follows Gleicher's visual comparison paper \cite{gleicher2017considerations}. 

% Human motion comparison \cite{bernard2017approaches}

% Note that this is different from visualization of time-warped data (e,g., \cite{solteszova2020memento})

% Szafirs' survey paper \cite{szafir2021connecting}

\begin{table*}[tb]
\caption{Motion Properties Being Compared}
\label{tab:needs}
\vspace{-5mm}
\begin{center}
\begin{tabular}{llcccl}
\hline
\rule{0pt}{1.1\normalbaselineskip}%
 &Properties & \makecell[c]{Cartesian \\ Space} & \makecell{Joint \\ Space} & Time & Example Source\\[1mm]
\hline
\rule{0pt}{1.1\normalbaselineskip}%
\multirow{4}{*}{\rotatebox[origin=c]{90}{Objective}}% 
%\rule{0pt}{1.1\normalbaselineskip}%
& Efficiency &  &  & \newmoon & ... other robot trajectories that perform the same task more  \textbf{efficiently}...\cite{gomes2019humanoid}  \\
 %\rule{0pt}{1.1\normalbaselineskip}%
& Collision-Free/Safeness & \newmoon &  &  & ... lead to \textbf{safer} and socially more acceptable robot trajectories. \cite{kucner2017enabling} \\
%\rule{0pt}{1.1\normalbaselineskip}%
& Smoothness/Continuity & \newmoon & \newmoon  & \newmoon &  ...  in terms of faster search, and \textbf{smoother} and shorter robot trajectories. \cite{guruprasad2010multi}   \\
%\rule{0pt}{1.1\normalbaselineskip}%
& Accuracy & \newmoon & & & 
 %\makecell[l]{ Our method achieves robot motion \textbf{accuracies} that outperform the baseline. \cite{zhao2018collaborative}}\\
... can be used to automatically generate \textbf{accurate} 6 DOF robot paths... \cite{zhang2006line} \\
\hline 
\rule{0pt}{1.1\normalbaselineskip}%
\multirow{3}{*}{\rotatebox[origin=c]{90}{\makecell{Subjec-\\tive}}}%
%\rule{0pt}{1.1\normalbaselineskip}%
&Predictability & \newmoon & & \newmoon &  % \makecell[l]{... the robot’s motion will become significantly more \textbf{predictable}. \cite{dragan2014familiarization}}  \\
 
 \makecell[l]{Robot motion should also be ... when operating among people, \textbf{predictable}. \cite{phillips2013anytime}} \\
 
%\rule{0pt}{1.1\normalbaselineskip}%
&Legibility & \newmoon & & \newmoon & \makecell[l]{... thus allowing the generation of smoother and more \textbf{legible} robot motion. \cite{palmieri2021guest}} \\
%\rule{0pt}{1.1\normalbaselineskip}%
&Human-Like/Natural & \newmoon &   & \newmoon &  \makecell[l]{ We demonstrate a method for making robot motion more \textbf{human-like}. \cite{gielniak2011spatiotemporal}} \\
 
 % Socially acceptable  \\
 % Stable \\
\hline
\vspace{-10mm}
\end{tabular}
\end{center}
\end{table*}

\section{Problem Overview} \label{sec:overview}
In the scope of this paper, we define a \textit{robot motion} as a collection of motion data that encodes how a task is performed. A robot motion may contain the movements of multiple robots and relevant objects,  \textit{e.g.}, the object being manipulated. 
The motion of a mobile robot, a drone, or an object, is represented by a trajectory of poses: $\chi: [0,T] \rightarrow SE(3)$.  The motion of a high-degree-of-freedom robot manipulator or humanoid robot is encoded using a trajectory of joint states:  $\xi: [0,T] \rightarrow \mathbb{R}^n$, where $n$ is the number of joints.
Therefore, robot motions are time-varying, high-dimensional data. This paper aims to provide visualization approaches to facilitate the understanding and comparison of a small number of robot motions. \added{The visual comparison of a large number of robot motions presents unique challenges and likely requires different approaches.} Visualization approaches include a collection of 2D graphs, 3D shapes, interactive tools, layouts, and mechanisms such as multi-view coordination. 
With multiple visualizations and flexible layouts, roboticists can create workflows that support their needs.

\vspace{-3mm}
\section{Design Process} \label{sec:process}
We develop Motion Comparator by applying a rigorous design framework \cite{gleicher2017considerations} which identifies roboticist needs, analyzes comparative challenges and strategies, and considers different visual designs. This section describes our design process and the design choices we make. 

\subsection{Comparative Needs} \label{sec: needs}

% \subsection{Properties Being Compared} \label{sec:properties}
\replaced{W}{In order to design a visual comparison tool, w}e first identify the properties of robot motions that are of interest to roboticists. Robot motions are compared by their similarities or differences in these properties. 

We systematically identified comparative properties of robot motions from the abstracts of 64,893 papers published in major robotics journals (\textit{e.g.}, the International Journal of Robotics Research), and robotics conferences (\textit{e.g.}, Robotics: Science and Systems). The first step was a filtering process to identify sentences that contain ``robot motion'', ``robot trajectory'', ``robot path'', ``robot movement'', and their respective plural forms. 
We collected adjectives and adverbs among the identified sentences.
After manually removing ambiguous words (\textit{e.g.}, ``important'') and combining synonyms, we further examined the remaining words in their original sentences to ensure that they were used to describe robot motions.

Our process identified a total of seven properties. We report the properties, along with their representative sources, in Table \ref{tab:needs}. Some of the identified properties are considered \textit{objective}, \textit{i.e.}, these properties can be unambiguously described in mathematics and communicated using certain metrics. Meanwhile, it is more challenging to measure and communicate \textit{subjective} properties because their criteria are set by individuals and may change depending on the contexts. Motion Comparator aims to facilitate the assessment of both objective and subjective properties of robot motions. 

The identified properties describe robot motions from different perspectives. 
Because robot motions are time-varying, high-dimensional data, we seek to \textit{project} robot motions to some lower dimensional space to reveal the identified properties. 
As shown in Table \ref{tab:needs}, the identified properties describe motions in three spaces: the Cartesian space which encompasses both robots and humans, the joint space which unambiguously describes a robot configuration, and time which demotes the duration of motions. In order to \replaced{cover all properties in the spaces}{sufficiently visualize the identified properties}, Motion Comparator provides \replaced{3D scenes, time-series plot, and timeline bar to}{various views that} display motions in these spaces. We will describe these views in \cref{sec:views}.

\subsection{Comparative Challenges and Strategies} \label{sec:challenges}
Gleicher \cite{gleicher2017considerations} classifies the challenges in visual comparison into three categories: the large number of items, the complexity of individual items, and the complexity of the relationships between items. While robot motion comparison has all these challenges, this paper focuses on the visual comparison of a small number of robot motions, avoiding the first type of challenge. 
We identify the remaining challenges as the temporal length and spatial complexity of each robot motion, as well as the complex relationships between robot motions. Below, we describe the three challenges, our strategies to address them, and the corresponding features offered by Motion Comparator.  

\textit{Challenge 1: Temporal length of robot motions}  --
Robot motions are time-varying data. 3D scene visualizations only show the robot state at a specific timestamp and users need to use the animated timeline bar to view an entire motion. New features should be provided to effectively handle the temporal length of robot motions. 

\textit{Strategy 1.1: Traces in various spaces}  -- 
A \textit{trace} is a line that displays the positions of an object as time changes. Traces address the temporal length challenge of robot motions by folding time into space. 
Depending on the space being displayed, Motion Comparator offers traces in Cartesian, quaternion, and joint spaces. Position traces are lines in Cartesian space that depict an object's \textit{position} changes over time. Quaternion traces are lines in 4D Euclidean space that encode an object's \textit{orientation} changes. Joint space traces are lines in a robot's joint space that show how its \textit{joints} move over time. The implementation and visualization of these traces are described in \cref{sec:cartesian_trace}, \cref{sec:quaternion_trace}, and \cref{sec:joint_trace}.

\textit{Strategy 1.2: Subset selection}  --
Another strategy to visualize temporally long motions is to select a subset of the motions. Similar to video editor software, Motion Comparator features a timeline bar that includes a selectable region, allowing users to control the duration of the motion being visualized. This feature enables users to temporally ``zoom in''  to inspect a particular part of a motion. 
  
\textit{Challenge 2: Spatial complexity of robot motions}  --
At each time frame, a robot may move its joints in joint space, resulting in position and orientation changes in Cartesian space. 
A robot's motion in Cartesian space determines its task performance, \textit{e.g.}, accuracy and collision avoidance. 
On the other hand, the robot's energy usage, wear, and tear are influenced by its motion in joint space. Hence, it is necessary for roboticists to possess a comprehensive understanding of motions in both Cartesian and joint space.

\textit{Strategy 2: Multi-view coordination}  --
This strategy aims to assist users in building connections between various visualizations. Our tool provides a scrubbable timeline bar that enables users to examine each frame on robot motions. Multiple visualizations are controlled by the same timeline bar. 
When users scrub the timeline bar, the 3D scene will visualize the robot and its surrounding environment at the selected time stamp. Meanwhile, the corresponding points on quaternion traces, joint traces, and time-series lines will be highlighted (Fig. \ref{fig: teaser}). In addition, we offer synchronization of views and lighting across views. These features enable users to examine multiple aspects of a robot motion at once. 

\textit{Challenge 3: Complexity of relationships}  --
Robot motions are time-series data that may have different durations, so the direct comparison at each time frame may not be meaningful. For example, when comparing motions for a pick-and-place task, it is more meaningful to compare the robots' poses as they pick up the object, as opposed to their poses at a specific time frame. 
This example demonstrates the complex relationship between motions, urging comparison tools to seek ways to simplify the comparison. 

\textit{Strategy 3: Temporal alignments}  --
Given that motions are not necessarily frame-to-frame corresponded, our tool facilitates comparison by autonomous temporal alignment. Specifically, we use \textit{time-warping} \cite{senin2008dynamic} to establish a correspondence that minimizes some loss function. 
Motion Comparator offers loss functions as the difference of joint states in joint space, the Euclidean distance in Cartesian space, or a sum of multiple objectives. \deleted{While our current implementation uses the dynamic time warping algorithm \cite{senin2008dynamic}, we plan to integrate more advanced time warping methods to Motion Comparator, \textit{e.g.}, GDTW \cite{deriso2023general}.}

\subsection{Comparative Designs} \label{sec:design}
As a visual comparison tool, Motion Comparator must decide how information is displayed to assist comparison. Prior works \cite{gleicher2017considerations} have identified three basic categories for comparative designs: juxtaposition, superposition, and explicit encoding. As described in \cref{sec: motion_visualization}, these designs have their benefits and drawbacks. Motion Comparator incorporates all these three designs to satisfy various comparison needs. For instance, superposition designs are suitable for detecting subtle differences between robot motions while explicit encoding designs are beneficial for measuring or communicating a relationship. In \cref{sec:views}, we will further describe how Motion Comparator was built from the ground up to support comparative designs.

\vspace{-2mm}
\section{Multi-View Interface} \label{sec:views}

\begin{figure}[tb]
  \centering
  \includegraphics[width=3.45in]{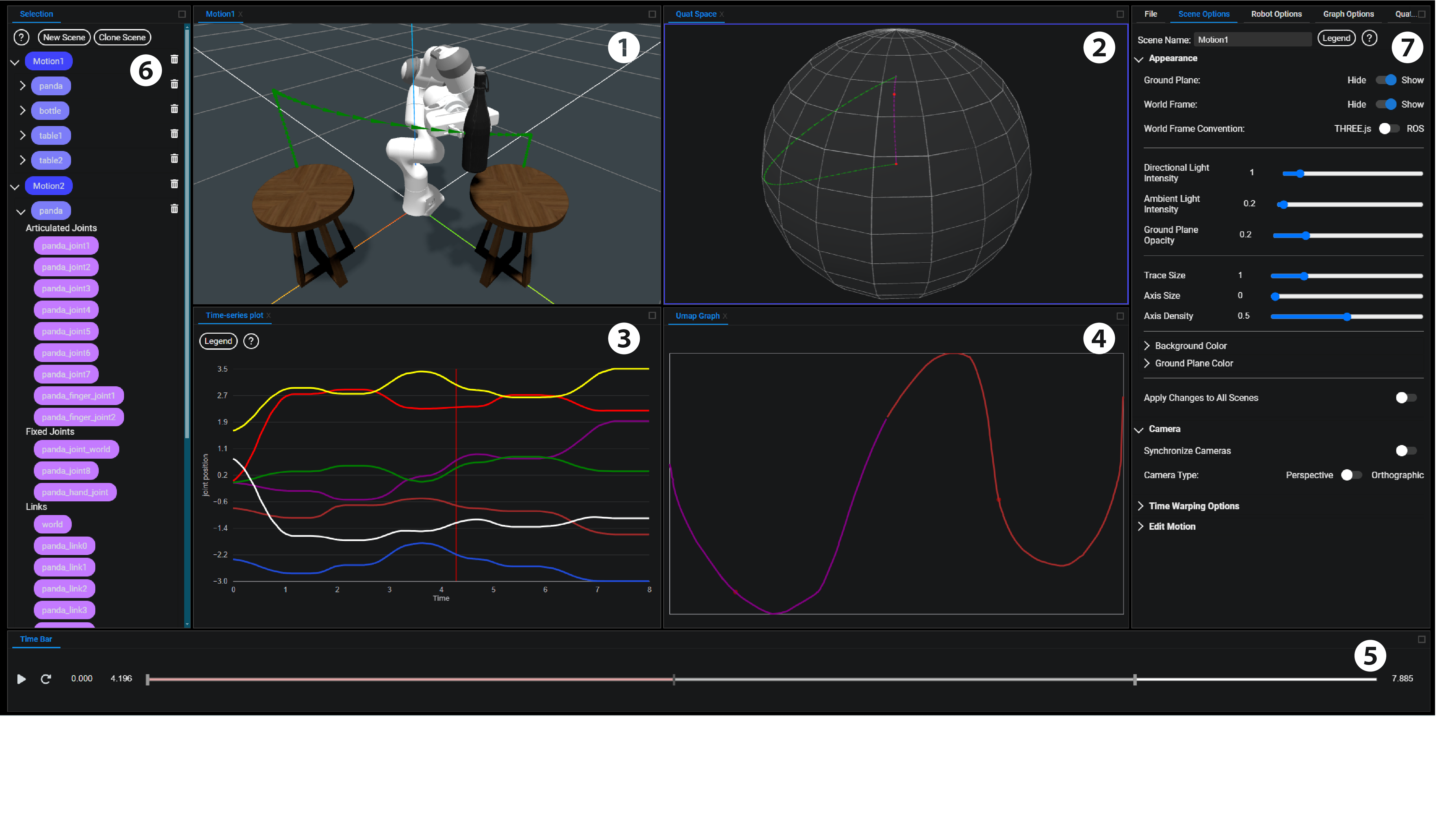}
  \vspace{-12mm}
  \caption{Screenshot of Motion Comparator showing its multiple panels. Its customizable layout allows users to tailor the interface for various workflows. \protect\circled{1} 3D Scene with a position trace \protect\circled{2} quaternion space \protect\circled{3} time-series plot \protect\circled{4} UMAP graph \protect\circled{5} timeline bar \protect\circled{6} motion library that shows loaded robot motions \protect\circled{7} option panels to change preferences. }
  \label{fig: teaser}
  \vspace{-5mm}
\end{figure}
\begin{figure*} [tb]
  \centering
  \includegraphics[width=7in]{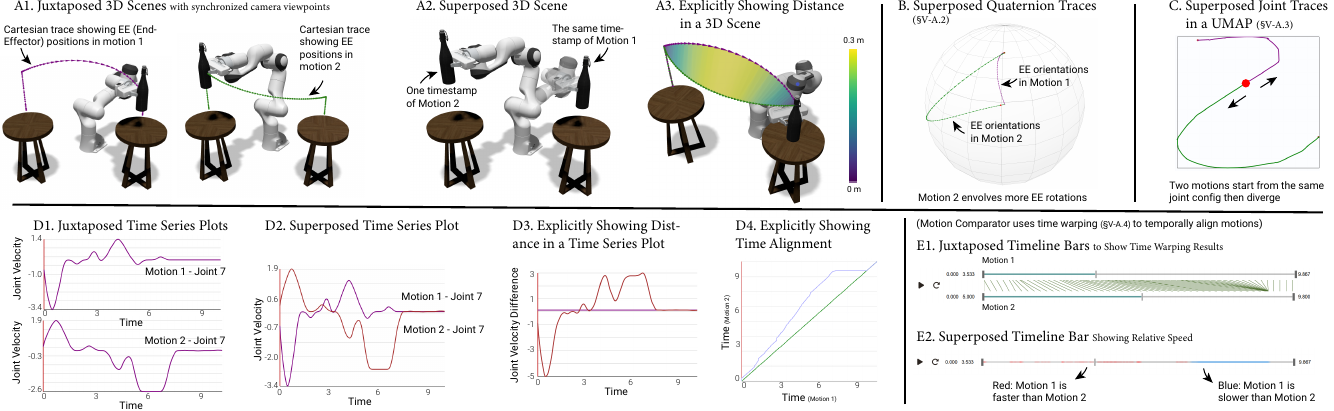}
  \vspace{-5mm}
  \caption{Motion Comparator offers five views to visualization robot motions, including 3D scene (A), quaternion space (B), UMAP graph (C), time-series plot (D), and timeline bar (E). These views incorporate the comparative strategies in \cref{sec:challenges}, including position trace (A1), quaternion trace (B), joint trace (C), and temporal alignments (D4, E1, E2). These views also incorporate comparative designs, including juxtaposition, superposition, and explicit encoding.  }
  \label{fig: 3x3}
  \vspace{-5mm}
\end{figure*}

Motion Comparator provides a multi-view interface that visualizes various aspects of motions to satisfy the comparative needs identified in \cref{sec: needs}. Although certain views may be present in some existing visualization tools, the views in Motion Comparator are built from the ground up to incorporate the strategies in \cref{sec:challenges} and the comparative designs in \cref{sec:design}. 
For example, roboticists can juxtapose two instances of Rviz side-by-side for visual comparison. However, users need to manually synchronize camera viewpoints between the instances and temporally align motions. \replaced{In contrast}{Meanwhile}, the 3D scenes of Motion Comparator offer features that assist comparison, such as unifying viewpoints and lighting conditions, as well as automatic temporal alignments. 
Below, we first explain the basic features of each view, then how it incorporates comparative strategies and designs. 

\subsection{3D Scene} \label{sec:3d_scene} \label{sec:cartesian_trace}
3D scenes in Motion Comparator render robots and their surroundings in simulated 3D environments. \replaced{T}{In order t}o navigate through 3D scenes, users can adjust the camera viewpoint using orbit controls with their mouse. Motion Comparator also offers configurability for customizing lighting and opacity, manually adjusting robot motions, and positioning objects in 3D scenes. 3D scene views are essential for users to see and understand how a robot moves. With an animated timeline bar, robot motions can be played in 3D scenes. 

As discussed in \cref{sec:challenges}-Strategy 1.1, \textit{position traces} fold time into space to handle temporally long motions. A position trace is a line in a 3D scene that depicts an object's position at each time stamp \cite{ware2006visualizing}. As shown in Fig. \ref{fig: 3x3}-A1, Motion Comparator draws position traces with consecutive cones to indicate directions and encode velocity.

The customizable layout feature (\cref{sec:implementation}-3) of Motion Comparator allows users to place multiple 3D scenes side by side, thereby achieving \textit{juxtaposition} designs\replaced{ (see Fig. \ref{fig: 3x3}-A1)}{. As shown in Fig. \ref{fig: 3x3}-A1, Motion Comparator offers the functionalities to synchronize camera viewpoints, lighting conditions, and background colors across all 3D scenes.}  
Motion Comparator can also visualize multiple robot motions in a single 3D scene, thereby achieving \textit{superposition} designs. As shown in Fig. \ref{fig: 3x3}-A2, users can change the opacities of robot models and objects to alleviate the problem of occlusion. 
In addition, Motion Comparator provides another approach that \textit{explicitly} visualizes distances between objects in Cartesian space. 
As shown in Fig. \ref{fig: 3x3}-A3, by filling the gap between Cartesian traces and using color to encode the various distances, Motion Comparator contextualizes the visualization of differences in 3D scenes.

\vspace{-1mm}
\subsection{Quaternion Space}  \label{sec:quaternion_trace}
As described in \cref{sec: motion_visualization}, it is challenging to visualize the orientation change of an object. 
One approach commonly used in data visualization represents orientations as unit quaternions $\mathbf{q} = (x,y,z,w) \in \mathbb{R}^4$ and visualizes quaternions by projecting them from 4D Euclidean space to a sphere in 3D space \cite{Hanson2006, milanko2023wuda}.
Specifically, we project quaternions to a 3D space where $w=0$. After projection, each quaternion becomes a 3D vector that is collinear with the rotational axis and its length is correlated with the angles of rotation.

While it may take some training to interpret quaternion traces, they effectively visualize orientation changes \cite{milanko2023wuda}.
The utility of quaternion traces becomes more apparent when multiple quaternion traces are drawn together. As shown in Fig. \ref{fig: 3x3}-B, longer quaternion traces indicate more rotations.

\subsection{Time-Series Plots} \label{sec:plots}
Motion Comparator offers time-series plots to visualize various scalars. These scalars include positions in joint space or Cartesian space and their derivatives (velocities, accelerations, and jerks). To handle noisy data, Motion Comparator provides filters to smooth time-series plots.  

Motion Comparator allows users to place multiple time-series plots side by side to make \textit{juxtaposition} designs (Fig. \cref{fig: 3x3}-D1). 
Meanwhile, time-series plots naturally accommodate \textit{superposition} designs because multiple lines can be drawn in a single plot (Fig. \ref{fig: 3x3}-D2). 
To \textit{explicitly} visualize the relationship between two variables, Motion Comparator can visualize their differences in a time-series plot (Fig. \ref{fig: 3x3}-D3). \replaced{It}{Motion Comparator} supports \deleted{the visualization of the} difference between robot joint positions, distance between objects, or velocity, acceleration, and jerk difference in Cartesian space or joint space. 
In addition, Motion Comparator can draw warping curves to \textit{explicitly} show the time alignment between two motions. As shown in Fig. \ref{fig: 3x3}-D4, a warping curve is displayed in a 2D plot where both the $x$ and $y$ axes represent time. When a warping curve is above the diagonal reference line, it indicates that the $x$-axis motion is faster than the $y$-axis motion.

\vspace{-2mm}
\subsection{Joint Trace} \label{sec:joint_trace}
\replaced{
While a time-series plot effectively visualizes the movement of individual joints, the display becomes cluttered when all joints of a robot are shown simultaneously on a single plot. To address this issue, we provide \textit{joint trace}, which leverages dimensionality reduction technologies to depict the movement of \textit{all} joints of a robot in a line.}{While time-series plots visualize the movement of individual joints, a joint trace is a line that depicts the movement of \textit{all} joints.} 
As shown in Fig. \ref{fig: 3x3}-C, we visualize joint traces using Uniform Manifold Approximation and Projection (UMAP) \cite{mcinnes2018umap}, a manifold learning technique for dimension reduction. The objective of UMAP is to construct a low-dimensional graph where similar joint states are close to each other and dissimilar joint states are far away from each other. 
Despite the information loss in dimension reduction, joint trace in UMAP graph is an efficient way for roboticists to quickly identify patterns and gain intuitive insights (\cref{sec:param_tuning} provides an example). \added{Our current implementation uses the \texttt{umap-js} library for dimensionality reduction, which slows down with long motions. To improve performance, we plan to switch to a Python backend using the \texttt{umap-learn} library.
}

\subsection{Scrubbable Timeline Bar} \label{sec:timebar}
The timeline bar of Motion Comparator serves the purpose of indicating the duration of motions, animating robot motions, and providing users with a convenient way to navigate through time. In a manner akin to the utilization of timeline bars in video players, users can interact with the timeline bar by either clicking on it to navigate to a specific timestamp or dragging its scrubber to make fine adjustments. 

In addition, timeline bars visualize time-warping results and show the temporal alignment between motions. Fig. \ref{fig: 3x3}-E1) provides an example of the \textit{juxtaposition} design that compares the temporal characteristics of two motions.
When users move the scrubber on one timeline bar, the scrubber on the other timeline bar is automatically moved to the corresponding position. The lines between the timeline bars also illustrate the temporal alignment between motions.
Motion Comparator also visualizes time-warping results in a superposition design. 
As shown in Figure \ref{fig: 3x3}-E2, the timeline bar uses color to encode the relative speed between two motions, where blue and red indicate that a motion has a lower and higher speed in comparison to the reference motion. This design is particularly useful for users to dissect the temporal difference between motions. For instance, users can quickly identify that it is the slow pick-up motion that causes a motion to be less efficient than another.

\vspace{-1mm}
\section{Implementation} \label{sec:implementation}

In this section, we describe the implementation details of Motion Comparator, specifically highlighting its role in facilitating \textit{communication} of robot motions.

\subsubsection{Web-based interface}
Our goal is to build a web-based interface that is lightweight and cross-platform, eliminating the need to download or install software. Motion Comparator is a \deleted{static} web application using \texttt{React.js}. It uses \texttt{three.js} to implement 3D scenes and quaternion spaces, and \texttt{D3.js} for time-series plots and UMAP graphs. \deleted{It can be hosted on static web servers such as Github Pages.} \added{Motion Comparator is open-sourced on GitHub, allowing it to benefit from community feedback and facilitating ongoing maintenance.  }

\subsubsection{ROS integration}
Motion Comparator is built as a convenient tool for Robot Operating System (ROS) users. Motion Comparator can directly parse binary \texttt{rosbag} files that record robot motions. \added{In addition, Motion Comparator also reads motions from \texttt{csv} files, allowing users to visualize motions exported from other platforms. }

\subsubsection{Customizable layouts} \label{sec:customizable_layout}
As shown in Fig. \ref{fig: teaser}, Motion Comparator is multi-view application with several panels. The layout is customizable, \textit{i.e.},  allowing users to create, reposition, stack, or destroy views.
The customizable layouts allow users to configure the tool to best address their needs, such as creating multiple coordinated views for visualization and various juxtaposition designs for comparison.

\subsubsection{Save, restore, and share}
In Motion Comparator, a workspace contains all changes made by users, including loaded motion data, created visualizations, and customized layouts. A workspace can be saved to a file and shared by a link. By clicking on the link, a roboticist can restore the exact same workspace, seeing the same robot motion data in the same visualizations using the same layout\added{, thereby facilitating the communication of robot motions}.

\vspace{-1mm}
\section{\replaced{Validation}{Case Studies}} \label{sec:case_studies}

\replaced{Motion Comaraptor is designed for roboticists to compare robot motions. For this kind of visualization systems that support solving real-world problems faced by domain experts, the most common form of validation are case studies with real users, real problems, and real data \cite{sedlmair2012design}. In this section, }{In order to illustrate how Motion Comparator facilitates the visualization, comparison, and communication of robot motions,} we present four case studies in which Motion Comparator is utilized for motion selection, troubleshooting, parameter tuning, or motion review. \added{Additionally, we invited two roboticists to use Motion Comparator to perform the tasks in the case studies. Both roboticists flexibly used features in our tool, including position traces, quaternion traces, UMAP graphs, and juxtaposition designs to compare motions. Their insights and feedback were collected and incorporated into our system to enhance its usability. In addition, we also used our tool as an educational tool to explain basic concepts of robot motions to a novice robotics student. }
All the case studies can be interactively explored using our online demo page\footnote{Demo page: \url{https://pages.graphics.cs.wisc.edu/MotionComparator}}.

\begin{figure}[tb]
  \centering
  \includegraphics[width=3.45in]{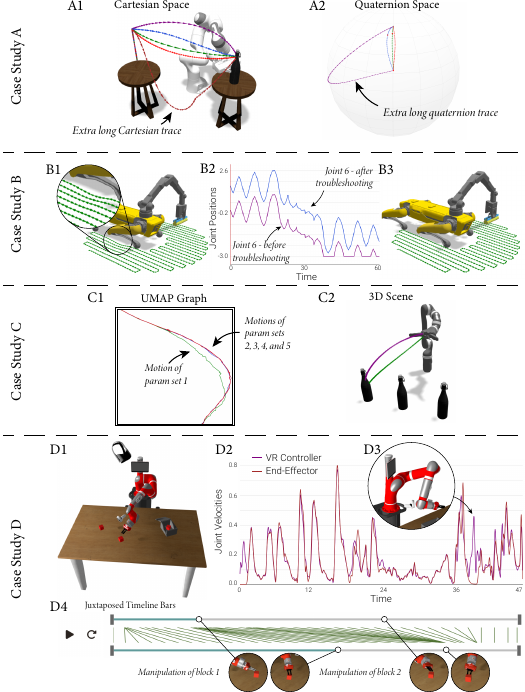}
  \vspace{-7mm}
  \caption{ Visualizations for case studies in \cref{sec:case_studies}.}
  \label{fig: demos}
  \vspace{-7mm}
\end{figure}

\subsection{Motion Selection after Planning} \label{sec:motion_selection}
Many common planning algorithms involve randomization. It is a widely used approach to generate multiple motion candidates using one or more motion planners, and subsequently select one motion to deploy to the physical robot. 
%This highlights the importance of motion comparison. 
Suppose a roboticist uses a sampling-based motion planner, Transition-based Rapidly-exploring Random Trees \cite{devaurs2013enhancing}, to generate motions for a Franka Emika Panda robot to pick up a water bottle and place it elsewhere. The randomness of the sampling-based motion planner means that it generates a different motion at each run. Without Motion Comparator, \replaced{the standard approach typically requires the roboticist to}{the roboticist may} play each motion one by one in a visualization tool such as Rviz. The roboticist needs to memorize all the motions and mentally compare them. Using Motion Comparator, the position traces first exclude one motion because it involves redundant movements (Figure \ref{fig: demos}-A1). 
Afterward, another motion is disregarded because of its significantly longer quaternion trace (Figure \ref{fig: demos}-A2). Longer quaternion traces indicate more rotation is applied to the bottle. Finally, the 3D scenes of Motion Comparator enable the roboticist to examine the remaining three motions and identify the best motion to deploy.

\subsection{Troubleshooting} \label{sec:troubleshooting}

Despite the existence of sophisticated motion planning algorithms, roboticists still play a crucial role in determining many aspects of motion generation, including starting configurations, placements of a robot, and ways to attach end-effectors. These decisions directly impact the qualities of robot motions. Motion Comparator helps roboticists troubleshoot and improve motion qualities through various visualizations.
In this case study, a roboticist programs a Boston Dynamics Spot robot equipped with a robot arm for a floor-wiping task. After parking at certain locations,  the robot wipes nearby areas using the cleaning tool in hand. The arm follows a pre-defined lawn-mower path using an optimization-based inverse kinematics solver, \textit{RangedIK} \cite{wang2023rangedik}. There are multiple ways to attach the cleaning tool to the robot's hand and the roboticist chooses one according to their experience. The position trace in Motion Comparator indicates that the robot is not capable of constantly following the given path (Figure \ref{fig: demos}-B1). Without Motion Comparator, the roboticist may manually develop code to inspect joint positions, velocities, accelerations, or jerks. \replaced{With multiple visualization features being integrated into Motion Comparator, the roboticist can identify the problem in the 3D scene, navigate to the problematic segment using the timeline bar, and troubleshoot using a time-series plot. The time-series plot reveals}{Using Motion Comparator, the time-series plots serve as a convenient diagnostic tool, revealing} that joint 6 of the robot's arm is restricted by its lower position limit (Figure \ref{fig: demos}-B2). Knowing the cause, the roboticist chooses an alternative way to mount the tool. The robot is able to accurately track the sweeping path now that joint 6 is no longer limiting it (Figure \ref{fig: demos}-B3).

\subsection{Parameter Tuning for Legible Motions}  \label{sec:param_tuning}

Many motion generation methods involve parameters that need manual tuning, \textit{e.g.}, the weights in a weighted-sum multiple-objective optimization-based approach \cite{bodden2018flexible}, the hyperparameters in reinforcement learning-based approaches \cite{kober2013reinforcement}, and the heuristic parameters in sampling-based motion planners \cite{lindemann2005current}. In order to understand the ramifications of parameters, roboticists need to \textit{compare} the motions generated using different parameters.
\iffalse
\begin{figure}[tb]
  \centering
  \includegraphics[width=3.5in]{figures/parameter tuning.pdf}
  \vspace{-5mm}
  \caption{Visualizations for case study 3.}
  \label{fig: param_tuning}
  \vspace{-5mm}
\end{figure}
\fi
A roboticist utilizes an optimization-based motion generator \cite{bodden2018flexible}
%\footnote{\url{https://github.com/uwgraphics/trajectoryoptimizer-public}} 
to generate legible motions to pick up a bottle. Five sets of parameters generate five motions. Without Motion Comparator, the roboticist may initiate five instances of visualization tools (\textit{e.g.}, Rviz) to display the five motions side-by-side. The roboticist needs to manually synchronize camera viewpoints and temporally align the motions. Using Motion Comparator, the UMAP graph view reveals that the motions generated by four parameter sets are in close proximity (Figure \ref{fig: demos}-C1). As opposed to comparing all five motions, the roboticist simply compares two of them. Subsequently, the roboticist visually determines the legibility of motions using position traces (Figure \ref{fig: demos}-C2). Our tool enables the roboticist to identify the set of parameters that can produce the most legible motion.

\subsection{Review of Motions in Teleoperation} \label{motion_review}

Motions in the aforementioned case studies are generated by planning algorithms. Robot motions can also be generated through programming by demonstration approaches such as teleoperation \cite{wang2023exploiting}. In this case study, a roboticist develops a teleoperation system in which an operator utilizes a VR controller to teleoperate a Rethink Robotics Sawyer robot to pick up blocks and drop them in a box (Figure \ref{fig: demos}-D1). 
In order to inspect the system's performance, the roboticist conducts a comparative analysis between input human motions and output robot motions.
The time-series plot in Motion Comparator shows that the system has consistently small latencies (Figure \ref{fig: demos}-D2). 
The inconsistency in the plot helps the roboticist detect when the robot loses the capability to accurately mimic its operator due to singularities or self-collision (Figure \ref{fig: demos}-D3).
In order to gain insights into different operation strategies, \textit{e.g.}, the ways to pick up a block, the roboticist also needs to compare motions generated by various users. 
Without Motion Comparator, the roboticist may need to manually navigate through motions to observe how the robot picks up a block. Using Motion Comparator, the time warping feature offered by Motion Comparator enables the roboticist to quickly align two motions and compare various ways to pick up a block (Figure \ref{fig: demos}-D4).

\section{\replaced{Discussion and Conclusion}{Discussion}}
In this section, we discuss the limitations of our work and provide a conclusion of the paper.
\deleted{\\A. Limitations\\}
Our work has a number of limitations that highlight potential directions for future research. First, while our system supports visual comparison among multiple robot motions, its usability significantly declines with a large number of motions. Neither juxtaposition nor superposition designs scale to many objects \cite{gleicher2017considerations}. In addition, our time warping and explicit encoding designs are limited to two-way comparisons. Future work should explore the approaches to visualize and compare numerous robot motions. 
Second, Motion Comparator utilizes Uniform Manifold Approximation and Projection (UMAP) to reduce the high dimensionality of joint traces. Being a general dimension reduction tool, UMAP is not specifically designed for robot joint data and may produce discontinuous traces despite the input joint traces being continuous. Future research can explore novel dimension reduction approaches to reduce the dimensionality of robot joint data.
Third, while providing several visualization approaches, Motion Comparator places the burden on the user to select, combine and apply them to their problem. Future work should focus on the usability, learnability, and effectiveness of the approaches in supporting motion comparison. 
%\vspace{-2mm}
\deleted{\\B.Conclusion \\}
%\vspace{-1mm}

\added{Motion Comparator is a visualization tool that benefits roboticists by facilitating comprehension, comparison, and communication of robot motions. Our design process involves a literature-based requirement analysis, the adaptation of multiple visualization approaches for motion comparisons, and the application of multi-view coordination approaches to integrate these visualizations. We believe that our tool and the design ideas it embodies contribute to robotic research by directly facilitating motion comparison and informing the development of future robot visualization tools.}

\deleted{We designed Motion Comparator following a rigorous visualization design process.
The resulting system utilizes many visualization approaches to facilitate the understanding of motions, including 3D scenes, time-series plots, scrubbable timeline bars with multi-view coordination, and traces in either Cartesian, quaternion, or joint space. To enhance the comparison of robot motions, Motion Comparator utilizes time warping to temporally align motions and provides juxtaposition, superposition, and explicit encoding designs for various comparison needs. Motion Comparator facilitates motion communication by allowing roboticists to share motion data, as well as visualizations and layouts, with their peers.  We believe that our system is beneficial to roboticists by facilitating comprehension, comparison, and communication of robot motions.}

% \hlpink{Taken together, this work presents Motion Comparator, a web-based visualization application that enables roboticists to understand, compare, and communicate robot motions.} 
% Motion Comparator is designed for roboticists to playback, understand, compare, and communicate robot motions. We believe that our visualization tool is beneficial to robot programming in many ways, including parameter tuning during motion generation, motion selection prior to deployment, motion inspection after experiments, and communication of experiment outcomes. 

\vspace{-2mm}

\bibliography{reference}
\bibliographystyle{IEEEtran}

\end{document}